\pdfoutput=1

\documentclass[11pt]{article}

\usepackage[]{naacl2021}

\usepackage{times}
\usepackage{latexsym}
\usepackage{verbatim}
\usepackage{graphicx}
\usepackage{booktabs}
\usepackage{xcolor}
\definecolor{purple}{RGB}{75,0,130}
\definecolor{blue}{RGB}{51,153, 255}
\usepackage{footmisc}
\usepackage{lscape}
\usepackage{amssymb}
\usepackage{pifont}
\newcommand{\cmark}{\ding{51}}%
\newcommand{\xmark}{\ding{55}}%

\usepackage{tikz-cd}

\usepackage[T1]{fontenc}

\usepackage[utf8]{inputenc}

\usepackage{microtype}
\usepackage{enumitem}
\usepackage{hyperref}

\title{\texttt{CLEVR\_HYP}: A Challenge Dataset and Baselines for \\ Visual Question Answering with Hypothetical Actions over Images}

\author{Shailaja Keyur Sampat\thanks{$\frac{}{}$ corresponding author}, Akshay Kumar, Yezhou Yang and Chitta Baral \\
        Arizona State Universiy, USA \\
        \texttt{\{ssampa17,akuma216,yz.yang, chitta\}@asu.edu}}
        
\begin{document}
\maketitle
\begin{abstract}
Most existing research on visual question answering (VQA) is limited to information explicitly present in an image or a video. In this paper, we take visual understanding to a higher level where systems are challenged to answer questions that involve mentally simulating the hypothetical consequences of performing specific actions in a given scenario. Towards that end, we formulate a vision-language question answering task based on the CLEVR \cite{johnson2017clevr} dataset. We
 then modify the best existing VQA methods and propose baseline solvers for this task. Finally, we motivate the development of better vision-language models by providing insights about the capability of diverse architectures to perform joint reasoning over image-text modality\footnote{Dataset setup scripts and code for baselines are made available at \hyperlink{link}{https://github.com/shailaja183/clevr\_hyp}. For additional details about the dataset creation process, refer supplementary material.}. 

\end{abstract}

\section{Introduction}
In 2014, Michael Jordan, in an interview \cite{ieee} said that ``{\em Deep learning is good at certain problems like image classification and identifying objects in the scene, but it struggles to talk about how those objects relate to each other, or how a person/robot would interact with those objects. For example, humans can deal with inferences about the scene: what if I sit down on that?, what if I put something on top of something? etc. There exists a range of problems that are far beyond the capability of today's machines.}"

While this interview was six years ago, and since then there has been a lot of progress in deep learning and its applications to visual understanding. Additionally, a large body of visual question answering (VQA) datasets \cite{VQA,ren2015exploring,
hudson2019gqa} have been compiled and many models have been developed over them, but the above mentioned ``inferences about the scene'' issue stated by Jordan remains largely unaddressed. 

In most existing VQA datasets, scene understanding is holistic and questions are centered around information explicitly present in the image (i.e. objects, attributes and actions). As a result, advanced object detection and scene graph techniques have been quite successful in achieving good performance over these datasets. However, provided an image, humans can speculate a wide range of implicit information. For example, the purpose of various objects in a scene, speculation about events that might have happened before, consider numerous imaginary situations and predicting possible future outcomes, intentions of a subject to perform particular actions, and many more. 

 \begin{figure}
  \center
  \includegraphics[width=\linewidth]{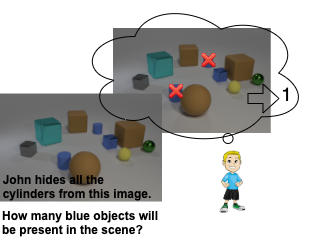}
  \caption{Motivation for the proposed \texttt{CLEVR\_HYP} dataset: an example demonstrating how humans can do mental simulations and reason over resulting scenario.}
  \label{fig:motivation}
\end{figure}

Among the above, an ability to imagine taking specific actions and simulating probable results without actually acting or experiencing 
is an important aspect of human cognition (Figure \ref{fig:motivation} gives an example of this). Thus, we believe that having autonomous systems equipped with a similar capability will further advance AI research.  This is particularly useful for robots performing on-demand tasks in safety-critical situations or navigating through dynamic environments, where they imagine possible outcomes for various situations without executing instructions directly.

Motivated by the above, we propose a challenge that attempts to bridge the gap between state-of-the-art AI and human-level cognition. The main contributions of this paper\footnote{Our work focuses on the capability of neural models to reason about the effects of actions given a visual-linguistic context and not on models that deal with intuitive physics.} are as follows;
\begin{itemize}
\itemsep0em \item We formalize a novel question answering task with respect to a hypothetical state of the world (in a visual form) when some action (described in a textual form) is performed.
    \item We create a large-scale dataset for this task, and refer it as \texttt{CLEVR\_HYP} i.e. VQA with hypothetical actions performed over images in CLEVR \cite{johnson2017clevr} style.
    \item We first evaluate the direct extensions of top VQA and NLQA (Natural language QA) solvers on this dataset. Then, we propose new baselines to solve \texttt{CLEVR\_HYP} and report their results.
    \item Through analysis and ablations, we provide insights about the capability of diverse architectures to perform joint reasoning over image-text modality. 
\end{itemize}

\begin{figure*}
            \begin{minipage}{0.2\linewidth}
            \textbf{I}: 
             \includegraphics[width=\linewidth,height=0.7\linewidth]{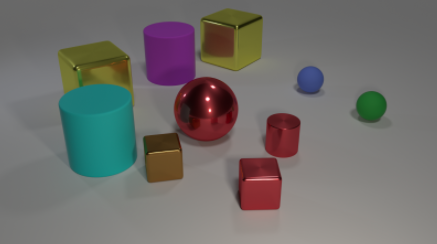}
            \end{minipage}
            \begin{minipage}{0.79\linewidth}
                 \begin{enumerate}
  \itemsep0em 
  \small{
     \item \textbf{T$_{A}$}: \textcolor{blue}{Paint} the small green ball with cyan color. \\
\textbf{Q$_{H}$}: Are there \textcolor{orange}{equal} yellow cubes on left of purple object and cyan spheres? (A: yes) 
    \item \textbf{T$_{A}$}: \textcolor{blue}{Add} a brown rubber cube behind the blue sphere that inherits its size from the green object.  \\
\textbf{Q$_{H}$}: \textcolor{orange}{How many} things are \textcolor{orange}{either} brown \textcolor{orange}{or} small? 
(A: 6)
    \item \textbf{T$_{A}$}: John \textcolor{blue}{moves} the small red cylinder on the large cube that is to the right of purple cylinder. \\
\textbf{Q$_{H}$}: \textcolor{orange}{What color} is the object that is at the bottom of the small red cylinder? 
(A: yellow) 
     }
  \end{enumerate}
            \end{minipage}
\caption{Three examples from \texttt{CLEVR\_HYP} dataset: given image (I), action text (T$_{A}$), question about hypothetical scenario (Q$_{H}$) and corresponding answer (A). The task is to understand possible perturbations in I with respect to various \textcolor{blue}{action(s)} performed as described in T$_{A}$. Questions test various \textcolor{orange}{reasoning capabilities} of a model with respect to the results of those action(s).} 
  \label{fig:examples}
\end{figure*}

\section{Related Work}





In this section we situate and compare our work with related areas such as implicit text generation/retrieval for a visual, visual question answering (VQA) over synthetic images, question answering (QA) involving hypothetical reasoning, and language-based manipulation in visual domains closest to \texttt{CLEVR\_HYP}. 

\paragraph{\textbf{Implicit Text Generation for a Visual:}} VisualComet \cite{park2020visualcomet} and Video2Commonsense \cite{fang2020video2commonsense} have made initial attempts to derive implicit information about images/videos contrary to traditional factual descriptions which leverage only visual attributes. VisualComet aims to generate commonsense inferences about events that could have happened before, events that can happen after and people's intents at present for each subject in a given image. They use a vision-language transformer that takes a sequence of inputs (image, event, place, inference) and train a model to predict inference in a language-model style. Video2Commonsense focuses on generating video descriptions that can incorporate commonsense facts related to intentions, effects, and implicit attributes about actions being performed by a subject. They extract top-ranked commonsense texts from the Atomic dataset and modify training objective to incorporate this information.  

While both involve a visual-textual component and actions, their key focus is about generating plausible events and commonsense respectively. Whereas, our work is related to performing certain actions and reasoning about its effect on the overall visual scene.   

\paragraph{Language-based Manipulation in Visual Domain: } Learning a mapping from natural language instructions to a sequences of actions to be performed in a visual environment is a common task in robotics \cite{kanu2020following,gaddy2019prelearning,ALFRED20}. 
Another relevant task is vision-and-language navigation \cite{mattersim,chen2018touchdown,nguyen2019vnla}, where an agent navigates in a visual environment to find goal location by following natural language instructions. Both above works include visuals, natural language instructions and a set of actions that can be performed to achieve desired goals. In this way, it is similar to our \texttt{CLEVR\_HYP}, but in our case, models require reasoning about the effect of actions performed rather than determining which action to perform. Also, we frame this in a QA style evaluation rather than producing instructions for low-level controls.  

Manipulation of natural images with language is an emerging research direction in computer vision. \cite{teney2020learning} proposed a method for generating counterfactual of VQA samples using image in-painting and masking. Also, there are works \cite{dong2017semantic, nam2018text, reed2016generative} which use Generative Adversarial Networks (GANs) \cite{goodfellow2014generative} for language conditioned image generation and manipulation. However, both the above tasks are more focused at object and attribute level manipulation rather than at action level.

\paragraph{\textbf{VQA over Synthetic Images:}} While natural images-based VQA datasets reflect challenges one can encounter in real-life situations, the requirement of costlier human annotations and vulnerability to biases are two major drawbacks. Contrary to them, synthetic datasets allow controlled data generation at scale  while being flexible to test specific reasoning skills. 

For the above reasons, following benchmark VQA datasets have incorporated synthetic images; COG \cite{yang2018dataset} and Shapes \cite{andreas2016neural} contain images with rendered 2D shapes; SHRDLU \cite{winograd1971procedures},
CLEVR \cite{johnson2017clevr}, and CLEVR-dialog \cite{kottur2019clevrdialog} have rendered scenes with 3D objects; DVQA \cite{kafle2018dvqa} and FigureQA \cite{kahou2017figureqa} have synthetically generated charts (bar chart, pie chart, dot-line etc.);  VQA-abstract \cite{VQA} and IQA \cite{gordon2018iqa} involves question-answering over synthetically rendered clipart-style scenes and interactive environments respectively. Our proposed dataset \texttt{CLEVR\_HYP} uses CLEVR \cite{johnson2017clevr} style rendered scenes with 3D objects as a visual component. It is distinct from all other synthetic VQA datasets for two key reasons; first, integration of action domain in synthetic VQA and second, the requirement of mental simulation in order to answer the question.     

\paragraph{\textbf{QA involving Hypothetical Reasoning:}} 
In the language domain, WIQA \cite{tandon2019wiqa} dataset tests the model's ability to do what-if reasoning over procedural text as a 3-way classification (the influence between pair of events as positive, negative or no-effect). In vision-language domains, a portion of TQA \cite{kembhavi2017you} and VCR \cite{zellers2019recognition} are relevant. Questions in TQA and VCR involve hypothetical scenarios about multi-modal science contexts and movie scenes respectively. However, none of the above two datasets' key focus is on the model's capability to imagine changes performed over the image.

\begin{figure}
  \center
  \includegraphics[width=\linewidth]{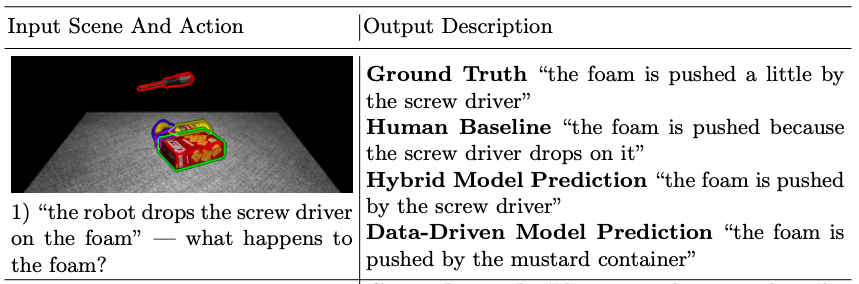}
  \caption{Example from TIWIQ \cite{wagner2018answering}.} 
  \label{fig:tiwiq}
\end{figure}

As shown in Figure \ref{fig:tiwiq}, the setting of TIWIQ (a benchmark dataset for ``physical intelligence'') \cite{wagner2018answering}  has some similarity with ours. It has synthetically rendered table-top scenes, four types of actions (push, rotate, remove and drop) being performed on an object and what-if questions. 

To our best knowledge, TIWIQ dataset is not publicly available. Based on our understanding from their manuscript, we observe following important distinction with this work. Our questions focus on the impact of actions on the whole image, while in TIWIQ questions are about impact of actions on a specific object in the image. Moreover, we frame \texttt{CLEVR\_HYP} as a classification task, contrary to TIWIQ which is a generative task. Our \texttt{CLEVR\_HYP} dataset has 175k automatically generated  image-action text-question samples which is much larger compared to TIWIQ which has only 1020 samples and manually crafted ground-truths. 

\section{\texttt{CLEVR\_HYP} Task and Dataset}
\label{sec:dataset}
Figure \ref{fig:examples} gives a glimpse of \texttt{CLEVR\_HYP} task. We opt for synthetic dataset creation as it allows automated and controlled data generation at scale with minimal biases. More details are described below.  

\paragraph{\textbf{3 Inputs:} Image(I), Action Text (T$_{A}$) and Hypothetical Question (Q$_{H}$)}
\paragraph{\textbf{1. Image(I):}} It is a given visual for our task. Each image in the dataset contains 4-10 randomly selected 3D objects rendered using Blender \cite{blender} in CLEVR \cite{johnson2017clevr} style. Objects have 4 attributes listed in the Table \ref{tab:attr}. Additionally, these objects can be referred using 5 relative spatial relations (left, right, in front, behind and on). We provide scene graphs\footnote{\label{note1}Scene graphs and Functional Programs (for action text and question) are not provided at the test-time.} containing all ground-truth information about a scene, that can be considered as a visual oracle for a given image.     
    \begin{table}[ht]
    \centering
\begin{tabular}{@{}ll@{}}
\cmidrule(r){1-2} 
\textbf{Attr.} & \textbf{Possible values in \texttt{CLEVR\_HYP}} \\ \cmidrule(r){1-2} 
\textbf{Color}     & gray, blue, brown, yellow,\\ & red, green, purple, cyan                                                         \\
\textbf{Shape}     & cylinder, sphere or cube                       \\
\textbf{Size}      & small or big                                                                                              \\
\textbf{Material}              & metal (shining) or rubber (matte)  
\\ \cmidrule(r){1-2} 
\end{tabular}
\caption{Object attributes in \texttt{CLEVR\_HYP} scenes.}
\label{tab:attr}
\end{table}

 \begin{figure*}
  \center
  (a) Function Catalog for \texttt{CLEVR\_HYP}, extended from CLEVR \cite{johnson2017clevr} \\
  \includegraphics[width=\textwidth]{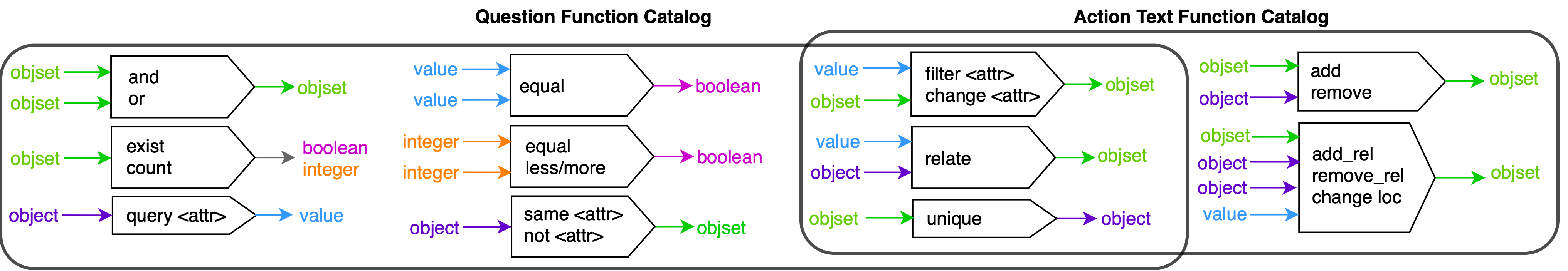}
  (b) Dataset creation pipeline \\ 
  \includegraphics[width=\textwidth]{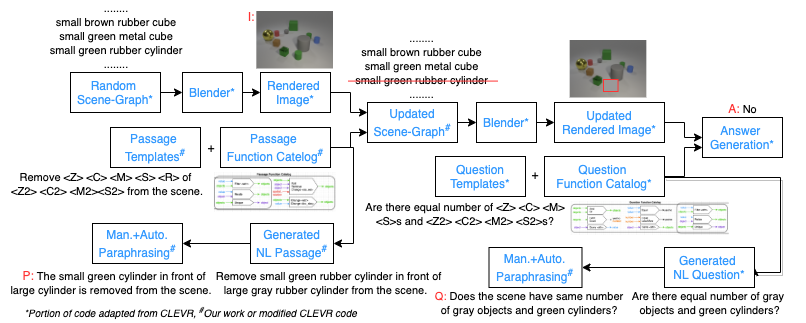}
  \caption{\texttt{CLEVR\_HYP} dataset creation process with example and function catalog used for ground-truth answer generation. (for more details, see Appendix \ref{sec:FC})}
  \label{fig:pipeline}
\end{figure*}

\paragraph{\textbf{2. Action Text (T$_{A}$):}} It is a natural language text describing various actions performed over the current scene. The action can be one of four: 
    \begin{enumerate}[label=(\roman*),noitemsep]       
        \item \textbf{Add} new object(s) to the scene
        \item \textbf{Remove} object(s) from the scene
        \item \textbf{Change} attributes of the object(s) 
        \item \textbf{Move} object(s) within scene (might be in plane i.e. left/right/front/back or out of plane i.e. move one object on top of another object\footnote{For simplicity, we assume that any object can be put on another object regardless of its size, material or shape.})
    \end{enumerate}
    
\noindent To generate action text, we start with manually written templates involving the aforementioned actions. For example, action involving change in the attribute of object(s) to a given value, we have a template of the following kind; \texttt{\small{`Change the $<$A$>$ of $<$Z$><$C$><$M$><$S$>$ to $<$V$>$'}}. Where $<$A$>$, $<$Z$>$, $<$C$>$,$<$M$>$,$<$S$>$,$<$V$>$ are placeholders for the attribute, size, color, material, shape and a value of attribute respectively.   
Each action text in the \texttt{CLEVR\_HYP} is associated with a functional program which if executed on an image’s scene graph, yields the new scene graph that simulates the effects of actions. 

Functional programs for action texts\textsuperscript{\ref{note1}} are built from the basic functions that correspond to elementary action operations (right part of Figure \ref{fig:pipeline}a). 
For the above mentioned `change' attribute action template, the equivalent functional program can be written as; \texttt{\small{`change\_attr($<$A$>$,filter\_size($<$Z$>$,filter \_color($<$C$>$, filter\_material($<$M$>$filter\_ shape($<$S$>$, scene())))),$<$V$>$)'}}. It essentially means, first filter out the objects with desired attributes and then update the value of their current attribute A to value V. 

\paragraph{\textbf{3. Question about Hypothetical Situation (Q$_{H}$):}} It is a natural language query that tests various visual reasoning abilities after simulating the effects of actions described in T$_{A}$. There are 5 possible reasoning types similar to CLEVR;
    \begin{enumerate}[label=(\roman*),noitemsep]   
        \item \textbf{Counting} objects fulfilling the condition
        \item \textbf{Verify existence} of certain objects
        \item \textbf{Query attribute} of a particular object
        \item \textbf{Compare attributes} of two objects 
        \item \textbf{Integer comparison} of two object sets  (same, larger or smaller)
    \end{enumerate}

\noindent Similar to action texts, we have templates and corresponding  programs for questions. Functional programs for questions\textsuperscript{\ref{note1}} are executed on the image’s updated scene graph (after incorporating effects of the action text) and yields the ground-truth answer to the question. Functional programs for questions are made of primitive functions shown in left part of the Figure \ref{fig:pipeline}a). \\

\noindent\textbf{Paraphrasing:} In order to create a challenging dataset from linguistic point of view and to prevent models from overfitting on templated representations, we leverage noun synonyms, object name paraphrasing and sentence-level paraphrasing. For noun synonyms, we use a pre-defined dictionary (such as cube\~block, sphere\~ball and so on). We programmatically generate all possibilities to refer to an object in the image (i.e. object name paraphrasing) and randomly sample one among them. For sentence level paraphrasing, we use Text-To-Text Transfer Transformer (T5) \cite{2020t5} fine-tuned over positive samples from Quora Question Pairs (QQP) dataset \cite{iyer2017first} for question paraphrasing. We use Fairseq \cite{ott2019fairseq} for action text paraphrasing which uses round-trip translation and mixture of experts \cite{shen2019mixture}. \\ 

\noindent Note that we keep the action text and question as separate inputs for the purpose of simplicity and keeping our focus on building solvers that can do mental simulation. One can create a simple template like ``$<$Q$_{H}$$>$ if $<$proper-noun/pronoun$>$ $<$T$_{A}$$>$?" or ``If $<$proper-noun/pronoun$>$ $<$T$_{A}$$>$, $<$Q$_{H}$$>$?" if they wish to process action and question as a single text input. For example, ``How many things are the same size as the cyan cylinder if I add a large brown rubber cube behind the blue object." or ``If I add a large brown rubber cube behind the blue object, how many things are the same size as the cyan cylinder?". However, having them together adds further complexity on the solver side as it first has to figure out what actions are performed and what is the question. 

By providing ground-truth object information (as a visual oracle) and machine-readable form of questions \& action texts (oracle for linguistic components). This information can be used to develop models which can process semi-structured  representations of image/text or for the explainability purposes (to precisely know which component of the model is failing).

\paragraph{\textbf{Output:}} \textbf{Answer (A)} to the Question (Q$_{H}$), which can be considered as a 27-way classification over attributes (8 colors + 3 shapes + 2 sizes + 2 material), numeric (0-9) and boolean (yes/no). \\

\begin{figure*}
  \center
  (a) Distribution based on Action Text types and Question types
  \includegraphics[width=\textwidth]{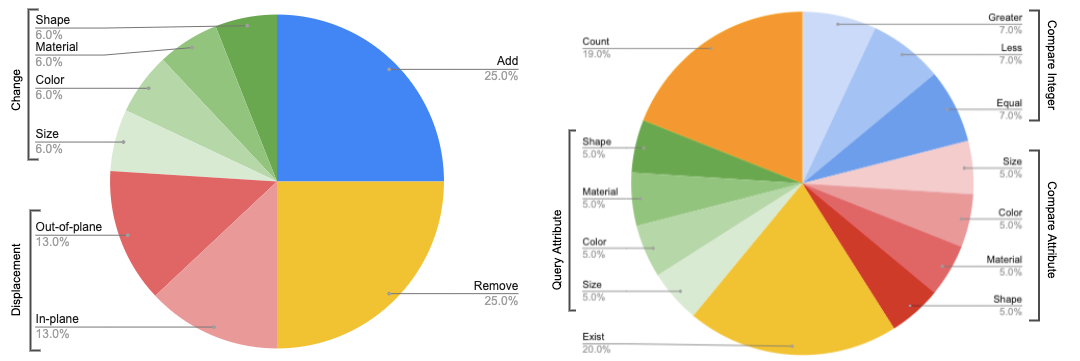}
  (b) Distribution of Answer types 
  \includegraphics[width=\textwidth]{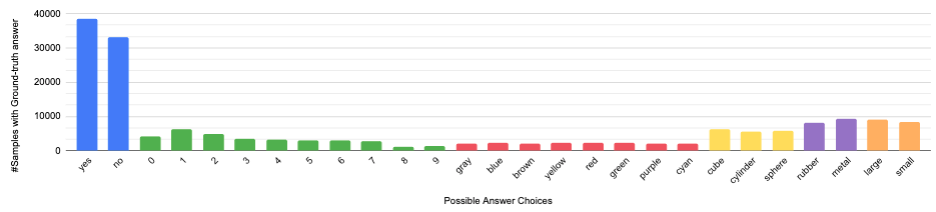}
  \caption{Visualization of distributions for actions, questions and answers in Original\_Train partition of \texttt{CLEVR\_HYP}.} 
  \label{fig:analysis}
\end{figure*}

\noindent\textbf{Dataset Partitions and Statistics:} We create \texttt{CLEVR\_HYP} dataset containing 175k image-action text-question samples using the process mentioned in Figure \ref{fig:pipeline}b. For each image, we generate 5 kinds of action texts (one for each add, remove, move in-plane and move out-of-plane and change attribute). For each action text type, we generate 5 questions (one for each count, exist, compare integer, query attribute and compare attribute). Hence, we get 5*5 unique action text-question pairs for each image, covering all actions and reasoning types in a balanced manner as shown in Figure \ref{fig:analysis}a (referred as Original partition). However, it leads to a skewed distribution of answers as observed from \ref{fig:analysis}b. Therefore, we curate a version of the dataset (referred as Balanced partition) consisting of 67.5k samples where all answer choices are equally-likely as well. \\ 
Additionally, we create two small challenge test sets (1500 image-action text-question samples each)- 2HopActionText (2HopT$_A$) and 2HopQuestion (2HopQ$_H$) to test generalization capability of the trained models. In 2HopT$_A$, we create action text which requires model to understand two different actions being taken on the scene. For example, `Add a small blue metal cylinder to the right of large yellow cube and remove the large cylinder from the scene.' and 'Move the purple object on top of small red cube then change its color to cyan.'. In 2HopQ$_H$, we create questions which require model to understand logical combinations of questions using `and', `or' and `not'. For example, `How many objects are either red or cylinder?' and `Are there any rubber cubes that are not green?'.

In Table \ref{tab:tab2}, we provide size of the various partitions and measure the diversity of the dataset in various aspects. For images, we calculate average number of objects present in the scene from the length of scene graph. For balanced partition, the number of images are much less compared to original, but more average number of objects per image. This is most likely due to the need to accommodate integers 4-9 more frequently as ground-truth answers. For textual components, we show average lengths (number of tokens separated by whitespaces) and count unique utterances as a measure of diversity. The original partition of the resulting dataset has   80\% and 83\% unique action text and questions respectively. For balanced partition, length and unique utterances for action text are nearly same as the original partition but for questions, it decreases. Questions in the original partition have been observed to enforce more strict and specific object references (such as small red metal cubes) compared to balanced partition (small cubes, red metal objects etc.), reducing the average length and uniqueness. It is intuitive for 2Hop partitions to have higher average length and uniqueness for $T_A$ and $Q_H$ respectively. This shows that despite having created this dataset from templates and rendered images with a limited set of attributes, it is still fairly challenging. 

 \begin{table*}
    \center
   \begin{tabular}{@{}lllllllll@{}}
\toprule
\textbf{Split} & \textbf{\#I} & \textbf{\begin{tabular}[c]{@{}l@{}}Avg.\\ \#Obj\end{tabular}} & \textbf{\#T$_{A}$} & \textbf{\begin{tabular}[c]{@{}l@{}}Unique\\ \#T$_{A}$\end{tabular}} &  \textbf{\begin{tabular}[c]{@{}l@{}}Avg.\\ $T_{A}$ Len.\end{tabular}} & \textbf{\#Q$_{H}$} & \textbf{\begin{tabular}[c]{@{}l@{}}Unique\\ \#Q$_{H}$\end{tabular}} &  \textbf{\begin{tabular}[c]{@{}l@{}}Avg.\\ Q$_{H}$ Len.\end{tabular}} \\ \midrule \midrule

 \textbf{Original\_Train}          &      5k   &  6.4 &      25k       &    20.7k                                                          &   12.8  &    125k     &    103.7k    & 22.6                                                     \\
\textbf{Original\_Val}           &     1k     &  6.7 &     5k        &     3.8k                  &   12.8                                       &         25k      &     20.9k          & 23.1                                             \\

 \textbf{Original\_Test}          &   1k  &  6.4      &      5k        &    3.6k      &   12.6                                                    &          25k   &  20.7k & 22.8    \\ 
 \midrule
\textbf{Balanced\_Train}          &   5k  &  7.6      &      25k        &    21.1k      &   12.8                                                    &          67.5k   &  58.2k & 20.3   \\ 
\textbf{Balanced\_Val}          &   1k  &  7.6      &      5k        &    3.9k      &   12.7                                                    &          13.5k   &  11.5k & 20.7    \\ 
\textbf{Balanced\_Test}          &   1k  &  7.5      &      5k        &    3.7k      &   12.6                                                    &          13.5k   &  11.4k & 20.4    \\ 
 \midrule                        
 \textbf{2Hop$T_A$\_Test}           &   1k        &  6.4 &      3k        &    2.6k  &    18.6                                                      &          15k   &  12.5k  & 22.8  \\ 
 
 \textbf{2Hop$Q_H$\_Test}      &   1k    &   6.4   &      3k        &    2.2k                                                          &  12.6  &        15k   &  13.7k  & 29.3 \\ \midrule \midrule

\end{tabular}
\caption{\texttt{CLEVR\_HYP} dataset splits and statistics (\# represents number of, k represents thousand).}
\label{tab:tab2}
\end{table*}

\section{Models that we experiment with}

Models trying to tackle \texttt{CLEVR\_HYP} dataset have to address four key challenges; 
    \begin{enumerate}[label=(\roman*),noitemsep]   
        \item understand hypothetical actions and questions in complex natural language,
        \item correctly disambiguate the objects of interest and obtain the structured representation (i.e. scene graphs or functional programs) of various modalities if required by the solver,
        \item understand the dynamics of the world based on the various actions performed over it,
        \item perform various kind of reasoning  to answer the question.  
       
    \end{enumerate}

\subsection{Random}
The QA task in \texttt{CLEVR\_HYP} dataset can be considered as a 27-class classification problem. Each answer choice is likely to be picked with a probability of 1/27. Therefore, the performance of the random baseline is 3.7\%.

\subsection{Human Performance}
We performed human evaluation with respect to 500 samples from the \texttt{CLEVR\_HYP} dataset. Accuracy of human evaluations on  original test, 2Hop$A_T$ and 2Hop$Q_H$ are 98.4\%, 96.2\% and 96.6\% respectively.

\subsection{Transformer Architectures}
Pre-trained transformer-based architectures have been observed \cite{li-etal-2020-bert} to capture a rich hierarchy of language-structures (text-only models) and effectively map entities/words with corresponding image regions (vision-language models). We experiment with various transformer-based models to understand their capability to understand the effects of actions on a visual domain.

\paragraph{Baseline 1- Machine Comprehension using RoBERTa:}
To evaluate the hypothetical VQA task through the text-only model, we convert images into the templated text using scene graphs. The templated text contains two kind of sentences; one describing properties of the objects i.e. ``There is a $<$Z$>$ $<$C$>$ $<$M$>$ $<$S$>$", the other one describing the relative spatial location i.e. ``The $<$Z$>$ $<$C$>$ $<$M$>$ $<$S$>$ is $<$R$>$ the $<$Z1$>$ $<$C1$>$ $<$M1$>$ $<$S1$>$". For example, ``There is a small green metal cube." and ``The large yellow rubber sphere is to the left of the small green metal cube". Then we concatenate templated text with the action text to create a reading comprehension passage. We use state-of-the-art machine comprehension baseline RoBERTa \cite{liu2019roberta} finetuned on the RACE dataset \cite{lai2017race}\footnote{architecture=roberta large, epochs=5, learning rate=$1\mathrm{e}{-05}$, batch size=2, update frequency=2, dropout=0.1, optimizer=adam with eps=$1\mathrm{e}{-06}$.}. Finally, we predict an answer to the question using this reading comprehension passage. 

\paragraph{Baseline 2- Visual Question Answering using LXMERT} 
Proposed by \cite{tan2019lxmert}, LXMERT is one of the best transformer based pre-trainable visual-linguistic representations which supports VQA as a downstream task. Typical VQA systems take an image and a language input. Therefore, to evaluate \texttt{CLEVR\_HYP} in VQA style, we concatenate action text and question to form a single text input. Since LXMERT is pre-trained on the natural images, we finetune it over \texttt{CLEVR\_HYP} dataset\footnote{epochs=4, learning rate=$5\mathrm{e}{-05}$, batch size=8} and then use it to predict answer.

\begin{figure*}
           \textbf{Nomenclature} I: Image, SG: Scene Graph, TT: Templated Text, $T_A$: Action Text, $Q_H$: Hypothetical Question, A: Answer, FP: Functional Program, ': Updated Modality  \\ \\
            \begin{minipage}{0.5\linewidth}
            \textbf{Baseline 1:\\}\\
            \begin{tikzcd}[cramped, sep=tiny]
            \\
            I \ar[r] 
            & SG \ar[r] 
            & TT + T_A \ar[dr]
            & \\
            &
            & & RoBERTa_{RACE} \ar[r] 
            & A \\
            &
            & \qquad Q_H \ar[ur]
            \end{tikzcd}
             \end{minipage}
            \begin{minipage}{0.5\linewidth}
            \textbf{Baseline 3:\\}\\
            \begin{tikzcd}[cramped, sep=tiny]
            & I \ar[dr] \\
            & & I' \ar[r] 
            & LXMERT_{CLEVR} \ar[r] 
            & A \\
            P \ar[r] 
            & FP \ar[ur] 
            & Q \ar[ur] 
            &
            \end{tikzcd}

            \end{minipage}
            \begin{minipage}{0.5\linewidth}
            \textbf{\\Baseline 2:\\}\\
             \begin{tikzcd}[cramped, sep=tiny]
            \\
            &\qquad  I \ar[dr] \\
            & & LXMERT_{CLEVR\_HYP} \ar[r] 
            & A \\
            & T_A + Q_H \ar[ur] 
            & 
            \end{tikzcd} 
            \end{minipage}
            \begin{minipage}{0.5\linewidth}
            \textbf{\\Baseline 4:\\}\\
            \begin{tikzcd}[cramped, sep=tiny]
            I \ar[r] 
            & SG \ar[dr] \\
            & & SG' & \longrightarrow
            &Symbolic \ar[r]
            & A \\
            P \ar[r] 
            & FP \ar[ur] 
            & Q \ar[r] 
            & FP \ar[ur] &
            \end{tikzcd}

            \end{minipage}
\caption{Graphical visualization of baseline models over \texttt{CLEVR\_HYP} described above.} 
  \label{fig:baselines}
\end{figure*}
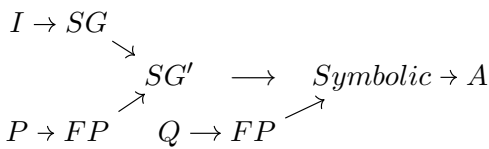

\subsection{Systematically incorporating effects of actions into neural models}
\paragraph{Baseline 3- Text-editing Image Baseline:}
In this method, we break-down the QA task with mental simulation in two parts; first, learn to generate an updated image (such that it has incorporated the effects of actions) and then perform visual question answering with respect to the updated image. We use the idea from Text Image Residual Gating proposed in  \cite{vo2019composing} to implement the first part. However there are two important distinctions; Their focus is on the retrieval from the given database. We modify their objective and develop text-adaptive encoder-decoder with residual connections to generate new image. Also, editing instructions in their CSS dataset \cite{vo2019composing} were quite simple. For example, `add red cube' and `remove yellow sphere'. In this case, one can add the red cube anywhere in the scene. We modify their architecture to precisely place objects to their relative spatial references (on left/right/front/ behind). Once we get the updated image, we feed it to the LXMERT \cite{tan2019lxmert} finetuned over the CLEVR \cite{johnson2017clevr} dataset along with the question and predict the answer.

\paragraph{Baseline 4- Scene Graph Update Model:} 
Instead of directly manipulating images, in this method, we leverage image scene graphs to convert image-editing problem into  graph-editing problem, conditioned on the action text. This is an emerging research direction to deal with changes in the visual modality over time or with new sources of information, as observed from recent parallel works \cite{chen2020graph, he2020scene}. 

We first use Mask R-CNN \cite{he2017mask} to get the segmentation mask of the objects and predict attributes (color, material, size, and shape) with an acceptance threshold of 0.9. Segmentation mask of each object along with original image is then passed through ResNet-34 \cite{he2016deep} to extract precise 3D coordinates of the object. We get the structured scene graph for the image. Then we use seq2seq with attention model originally proposed in \cite{johnson2017inferring} to generate functional programs (FP) for  action text and question. The execution engine executes programs on scene graph, implemented as a neural module network \cite{andreas2017neural} to update the scene representation and answer questions. 

We learn to update scene graphs according to functional program for the action text using reinforcement learning\footnote{finetuning learning rate=$1\mathrm{e}{-05}$, 1M iterations with early stopping, batch size=32}. The reward function is associated with our ground-truth program executor and generates reward if prediction exactly matches with ground-truth execution. Once we get the updated scene representation, we use neural-symbolic model\footnote{supervised pretraining learning rate=$7\mathrm{e}{-04}$, num iterations=20k, batch size=32 and then finetuning $1\mathrm{e}{-05}$, at most 2M iterations with early stopping, batch size=32} proposed by \cite{yi2018neural} to obtain the final answer. It is notable that \cite{yi2018neural} achieved near-perfect performance on the CLEVR QA task in addition to being fully explainable. 


\section{Baseline Results}
\label{sec:results}

In this section, we benchmark models described above on the \texttt{CLEVR\_HYP}. The dataset is formulated as a classification task with exactly one correct answer, so we use standard accuracy as evaluation metric. We then analyze their performance according to question and action types. 


\begin{table*}
\resizebox{\linewidth}{!}{%
\begin{tabular}{@{}llllllllllllllll@{}}
\toprule
\multicolumn{16}{c}{\textbf{Overall Baseline Performance for Various Test Sets of \texttt{CLEVR\_HYP} }}                                                                                                                                                                      \\ \midrule
\multicolumn{4}{c|}{\underline{\textit{Original Test}}}             & \multicolumn{4}{c|}{\underline{\textit{Balanced Test}}}             & \multicolumn{4}{c|}{\underline{\textit{2HopTA Test}}}            & \multicolumn{4}{c}{\underline{\textit{2HopQH Test}}} \\
BL1  & BL2  & BL3  & \multicolumn{1}{l|}{BL4}           & BL1  & BL2  & BL3           & \multicolumn{1}{l|}{BL4}  & BL1  & BL2  & BL3  & \multicolumn{1}{l|}{BL4}           & BL1    & BL2    & BL3   & BL4            \\
57.2 & 63.9 & 64.7 & \multicolumn{1}{l|}{\textbf{70.5}} & 55.3 & 65.2 & \textbf{69.5} & \multicolumn{1}{l|}{68.6} & 53.3 & 49.2 & 55.6 & \multicolumn{1}{l|}{\textbf{64.4}} & 55.2   & 52.9   & 58.7  & \textbf{66.5} \\
\bottomrule
\end{tabular}
}
$\frac{}{}$ \\

\resizebox{\linewidth}{!}{%
\begin{tabular}{@{}lllllllllllllll@{}}
\toprule
\multicolumn{15}{c}{\textbf{Performance break-down by Action Types and Reasoning Types for Baseline 3 and 4}}                                                                                                                                                                                                                                                    \\ \midrule
\multicolumn{1}{c}{\textbf{}} & \multicolumn{2}{c}{{\underline{ \textit{Original Test}}}} &  & \multicolumn{1}{c}{} & \multicolumn{2}{l}{{\underline{\textit{2Hop$A_T$ Test}}}} & \multicolumn{1}{l|}{} & \multicolumn{1}{c}{} & \multicolumn{2}{l}{{\underline{\textit{Original Test}}}} &  &     & \multicolumn{2}{l}{{\underline{\textit{2Hop$Q_H$ Test}}}} \\
                              & BL3                     & BL4                    &  &                      & BL3                    & BL4                   & \multicolumn{1}{l|}{} &                      & BL3                     & BL4                    &  &     & BL3                    & BL4                   \\
Add                           & 58.2                    & 65.9                   &  & Add+Remove           & 53.6                   & 63.2                  & \multicolumn{1}{l|}{} & Count                & 60.2                    & 74.3                   &  & And & 59.2                   & 67.1                  \\
Remove                        & 89.4                    & 88.6                   &  & Add+Change           & 55.4                   & 64.7                  & \multicolumn{1}{l|}{} & Exist                & 69.6                    & 72.6                   &  & Or  & 58.8                   & 67.4                  \\
Change                        & 88.7                    & 91.2                   &  & Add+Move             & 49.7                   & 57.5                  & \multicolumn{1}{l|}{} & CompInt              & 56.7                    & 67.3                   &  & Not & 58.1                   & 65.0                  \\
Move(in-plane)                & 61.5                    & 69.4                   &  & Remove+Change        & 82.1                   & 85.5                  & \multicolumn{1}{l|}{} & CompAttr             & 68.7                    & 70.5                   &  &     &                        &                       \\
Move(on)                      & 53.3                    & 66.1                   &  & Remove+Move          & 52.6                   & 66.4                  & \multicolumn{1}{l|}{} & QueryAttr            & 65.4                    & 68.1                   &  &     &                        &                       \\
                              &                         &                        &  & Change+Move          & 53.8                   & 63.3                  & \multicolumn{1}{l|}{} &                      &                         &                        &  &     &                        &                       \\ \bottomrule
\end{tabular}
}

\caption{Baseline performance over \texttt{CLEVR\_HYP} (BLx represents one of the four Baselines described above).}
\label{tab:tab3}
\end{table*}

Quantitative results from above experiments can be visualized in top part of the Table \ref{tab:tab3}. Among the methods described above, the scene graph update model has the best overall performance 70.5\% on original test data. Text-editing model is best over balanced set, but observed to have the poor generalization capability when two actions or reasoning capabilities have to be performed. \texttt{CLEVR\_HYP} requires models to reason about effect of hypothetical actions taken over images. LXMERT is not directly trained for this objective therefore, it struggles to do well on this task. The reason behind the poor performance of text-only baseline is due to its limitation to incorporate detailed spatial locations into the templates that we use to convert image into a machine comprehension passage. 

Two of our models (scene graph update and text-editing image) are transparent to visualize intermediate changes in the scene after performing actions. We analyse their ability to understand actions and make appropriate changes as shown in below part of Table  \ref{tab:tab3}. For the scene graph method, we compare the ground-truth functional program with the generated program and measure their exact-match accuracy. For the text-editing image method, we generate scene graphs for both images (original image and image after text-editing) and compare them. For attributes, we do exact-match, whereas for location information we consider matching only on the basis of relative spatial location. 

Both scene graph and text-editing models do quite well on `remove' and `change' actions whereas struggle when new objects are added or existing objects are moved around. The observation is consistent when multiple actions are combined. Therefore, actions remove+change can be performed with maximum accuracy whereas other combinations of actions accomplish relatively lower performance. It leads to the conclusion that understanding the effect of different actions are of varied complexity. Most models demonstrate better performance over counting, existence and attribute query type of questions than comparison questions. The scene graph update and text-editing methods show a performance drop of 6.1\% and 9.1\% respectively when multiple actions are performed on the scene. However, there is less of a performance gap for models on 2HopQ$_H$ compared to the test set, suggesting that models are able to better generalize with respect to multiple reasoning skills than complex actions. 


\section{Conclusion}
We introduce \texttt{CLEVR\_HYP}, a dataset to evaluate the ability of VQA systems after hypothetical actions are performed over the given image. We create this dataset by extending the data generation framework of CLEVR \cite{johnson2017clevr} that uses synthetically rendered images and templates for reasoning questions. Our dataset is challenging because rather than asking to reason about objects already present in the image, it asks about what would happen in an alternative world where changes have occurred. We provide ground-truth representations for images, hypothetical actions and questions to facilitate the development of models that systematically learn to reason about underlying process. We create several baseline models to benchmark \texttt{CLEVR\_HYP} and report their results. Our analysis shows that the models are able to perform reasonably well (70.5\%) on the limited number of actions and reasoning types, but struggle with complex scenarios. While neural models have achieved almost perfect performance on CLEVR and considering human performance as upperbound (98\%), there is a lot of room for improvement on \texttt{CLEVR\_HYP}. Our future work would include relaxing constraints by allowing a larger variety of actions, attributes and reasoning types. By extending this approach further for natural images, we aim to contribute in the development of better vision+language models. 



\section*{Acknowledgements}
We are thankful to the anonymous reviewers for the constructive feedback. This work is partially supported by the grants NSF 1816039, DARPA W911NF2020006 and ONR N00014-20-1-2332.

\bibliography{anthology,naacl2021}
\bibliographystyle{acl_natbib}

\newpage 
\appendix

\section{Appendix}
\label{sec:appendix}

\subsection{Relation of \texttt{CLEVR\_HYP} dataset with real-world situations}


Teaching methodologies leverage our ability to mentally simulate scenarios along with the metaphors to aid understanding about new concepts. In other words, to explain unfamiliar concepts, we often reference familiar concepts and provide additional clues to establish mapping between them. This way, a person can create a mental simulation about unfamiliar concept and aid basic understanding about it. 

\begin{figure}[h]
  (a) learning the concept `zebra' from the `horse' 
  \begin{center}
    \includegraphics[width=\linewidth]{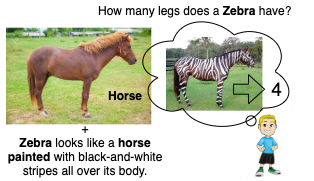}
    \end{center}
   (b) learning about `animal cell' by comparison with `plant cell' 
  \begin{center}
  \includegraphics[width=\linewidth]{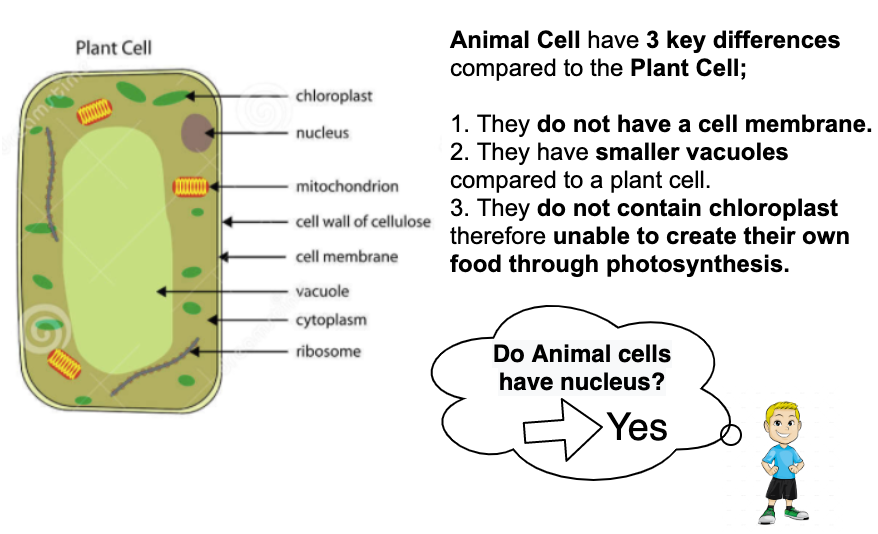}
  \end{center}
  \caption{Extension of \texttt{CLEVR\_HYP} for more complex real-world scenarios.} 
  \label{fig:future}
\end{figure}

For example, we want to explain a person how a `zebra' looks like, who has previously seen a `horse', we can do so using example in Figure \ref{fig:future}a. This naturally follows for more complex concepts. Let say, one wants to describe the structure of an atom to someone, he might use the analogy of a planetary system, where the components (planets $\sim$ electrons) circulate around a central entity (sun $\sim$ nucleus). One more such example is provided in Figure \ref{fig:future}b.

For humans, learning new concepts and performing mental simulations is omnipresent in day-to-day life. Therefore, \texttt{CLEVR\_HYP} dataset is very much grounded in the real world. Models developed on this dataset can serve a broad range of applications, particularly the ones where possible outcomes have to be predicted without actually executing the actions. For example, robots performing on-demand tasks in safety-critical situations or self-driving vehicles. In addition, these models can be an important component for other vision and language tasks such as automatic expansion of existing knowledge bases, zero shot learning and spatio-temporal visual reasoning.  

\subsection{Rejecting Bad Samples in \texttt{CLEVR\_HYP}}

Automated methods of question generation sometimes create invalid items, classified as `ill-posed' or `degenerate' by CLEVR \cite{johnson2017clevr} dataset generation framework. They consider question
``What color is the cube to the right of the sphere?" as ill-posed if there were many cubes right of the sphere, or degenerate if there is only one cube in the scene and reference to the sphere becomes unnecessary. In addition to this, we take one more step of quality control in order to prevent ordinary VQA models from succeeding over \texttt{CLEVR\_HYP} without proper reasoning.

\begin{figure}[h]
    \begin{center}
    \textbf{I}: \includegraphics[width=0.7\linewidth]{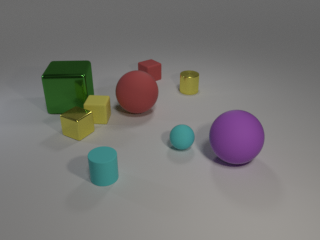} \end{center}
    Image-only model:  \\
    \textbf{Q1}: Is there any large sphere? \textbf{A}: \textcolor{blue}{Yes} \\
    \textbf{Q2}: Is there any large cube? \textbf{A}: \textcolor{orange}{Yes}\\ \\
    \texttt{CLEVR\_HYP}: \\
   \textbf{T}: Remove all matte objects from the scene. 
    \begin{center}\textbf{I'}:  \includegraphics[width=0.7\linewidth]{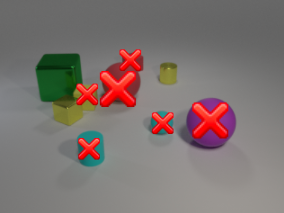} \end{center}
    \textbf{Q1}: Is there any large sphere? \textbf{A}: \textcolor{blue}{No} \cmark \\
    \textbf{Q2}: Is there any large cube? \textbf{A}: \textcolor{orange}{Yes} \xmark
    \caption{Validity of questions in \texttt{CLEVR\_HYP}}
    \label{fig:illquestions}
\end{figure}

In \texttt{CLEVR\_HYP}, one has to perform actions described in T over image I and then answer question Q with respect to the updated scenario. Therefore, to prevent ad-hoc models from exploiting biases in \texttt{CLEVR\_HYP}, we pose the requirement that a question must have different ground-truth answers for \texttt{CLEVR\_HYP} and image-only model. One such example is shown in Figure \ref{fig:illquestions}. For image (I), Q1 leads to different answers for CLEVR and \texttt{CLEVR\_HYP}, making sure that one needs to correctly incorporate the effect of T. Q2 is invalid for a given image-action text pair in the \texttt{CLEVR\_HYP} as one can answer it correctly without understanding T.

\subsection{More Examples from \texttt{CLEVR\_HYP}}

Beyond Figure \ref{fig:moreexamples}, all rest of the pages show more examples from our \texttt{CLEVR\_HYP} dataset. Each dataset item has 4 main components- image(I), action text (T$_A$), question about the hypothetical states (Q$_H$) and answer (A). We classify samples based on what actions are taken over the image and the kind of reasoning is required to answer questions.

\subsection{Function Catalog}
\label{sec:FC}
As described in Section \ref{sec:dataset} and shown in Figure \ref{fig:pipeline}, each action text and question is associated with a functional program. We provide more details about these basic functions in Table \ref{tab:fpcat} that was used to generate ground-truth answers for our dataset. Each function has input and output arguments, which are limited to following data types:
\begin{itemize}[noitemsep]
\item \textcolor{purple}{object}: a single object in the scene
    \item \textcolor{green}{objset}: a set of zero or more objects in scene
    \item \textcolor{orange}{integer}: an integer in [0,10]
    \item \textcolor{violet}{boolean}: `yes' or `no'
    \item \textcolor{blue}{values}: possible attribute values mentioned in Table \ref{tab:attr} 
\end{itemize}

\subsection{Paraphrasing}
In order to create a challenging dataset from the linguistic point of view and to prevent models from overfitting on templated representations, we leverage word synonyms and paraphrasing methods. This section provides more details about paraphrasing methods used in our dataset.

\begin{table*}
\begin{tabular}{@{}lll@{}}
\toprule
\textbf{Function}         & \textbf{Input Type → Output Type    }                       & \textbf{Return Value     }                                                                                                                                                    \\ \midrule
scene            & $\phi$ → \textcolor{green}{objset}                                      & Set of all objects in the scene                                                                                                                                      \\
unique           & \textcolor{green}{objset} → \textcolor{purple}{object}                                 & \begin{tabular}[c]{@{}l@{}}Object if objset is singleton; else raise exception\\ (to verify whether the input is unique or not)\end{tabular}               \\
relate           & \textcolor{purple}{object} × \textcolor{blue}{relation} → \textcolor{green}{objset}                      & Objects satisfying given spatial relation for input object                                                                                              \\
count            & \textcolor{green}{objset} → \textcolor{orange}{integer}                                & Size of the input set                                                                                                                                                \\
exist            & \textcolor{green}{objset} → \textcolor{violet}{boolean}                                & `Yes' if the input set is non-empty and `No' otherwise                                                                                                               \\
filter\_size     & \textcolor{green}{objset} × \textcolor{blue}{size} → \textcolor{green}{objset}                       & Subset of input objects that match the given size                                                                                                         \\
filter\_color    & \textcolor{green}{objset} × \textcolor{blue}{color} → \textcolor{green}{objset}                      & Subset of input objects that match the given color                                                                                                        \\
filter\_material & \textcolor{green}{objset} × \textcolor{blue}{material} → \textcolor{green}{objset}                   & Subset of input objects that match the given material                                                                                                     \\
filter\_shape    & \textcolor{green}{objset} × \textcolor{blue}{shape} → \textcolor{green}{objset}                      & Subset of input objects that match the given shape                                                                                                        \\
query\_size      & \textcolor{purple}{object} → \textcolor{blue}{size}                                      & Size of the input object                                                                                                                                  \\
query\_color     & \textcolor{purple}{object} → \textcolor{blue}{color}                                     & Color of the input object                                                                                                                                 \\
query\_material  & \textcolor{purple}{object} → \textcolor{blue}{material}                                  & Material of the input object                                                                                                                              \\
query\_shape     & \textcolor{purple}{object} → \textcolor{blue}{shape}                                     & Shape of the input object                                                                                                                                 \\
same\_size       & \textcolor{purple}{object} → \textcolor{green}{objset}                                 & Set of objects that have same size as input (excluded)                                                                                         \\
same\_color      & \textcolor{purple}{object} → \textcolor{green}{objset}                                 & Set of objects that have same color as input (excluded)                                                                                        \\
same\_material   & \textcolor{purple}{object} → \textcolor{green}{objset}                                 & Set of objects that have same material as input(excluded)                                                                                     \\
same\_shape      & \textcolor{purple}{object} → \textcolor{green}{objset}                                 & Set of objects that have same shape as input (excluded)                                                                                        \\

equal\_size      & \textcolor{blue}{size} × \textcolor{blue}{size} → \textcolor{violet}{boolean}                              & `Yes' if inputs are equal, `No' otherwise                                                                                                                     \\
equal\_color     & \textcolor{blue}{color} × \textcolor{blue}{color} → \textcolor{violet}{boolean}                     & `Yes' if inputs are equal, `No' otherwise                                                                                                                     \\
equal\_material  &      \textcolor{blue}{material} × \textcolor{blue}{material} → \textcolor{violet}{boolean}                        & `Yes' if inputs are equal, `No' otherwise                                                                                                                     \\
equal\_shape     & \textcolor{blue}{shape} × \textcolor{blue}{shape} → \textcolor{violet}{boolean}                            & `Yes' if inputs are equal, `No' otherwise                                                                                                                    \\
equal\_integer   & \textcolor{orange}{integer} × \textcolor{orange}{integer} → \textcolor{violet}{boolean}                        &  `Yes' if two integer inputs are equal, `No' otherwise                                                                                                         \\
less\_than       & \textcolor{orange}{integer} × \textcolor{orange}{integer} → \textcolor{violet}{boolean}                        & `Yes' if first integer is smaller than second, else `No'                                                                              \\
greater\_than    & \textcolor{orange}{integer} × \textcolor{orange}{integer} → \textcolor{violet}{boolean}                        & `Yes' if first integer is larger than second, else `No'  \\
and              & \textcolor{green}{objset} × \textcolor{green}{objset} → \textcolor{green}{objset}                  & Intersection of the two input sets                                                                                                                        \\
or               & \textcolor{green}{objset} × \textcolor{green}{objset} → \textcolor{green}{objset}                  & Union of the two input sets.   \\ \midrule \midrule
not\_size        & \textcolor{purple}{object} → \textcolor{green}{objset}                       & Subset of input objects that do not match given size                                                                                                  \\
not\_color       & \textcolor{purple}{object} → \textcolor{green}{objset}                      & Subset of input objects that do not match given color                                                                                                 \\
not\_material    & \textcolor{purple}{object} → \textcolor{green}{objset}                   & Subset of input objects that do not match given material                                                                                              \\
not\_shape       & \textcolor{purple}{object} → \textcolor{green}{objset}                      & Subset of input objects that do not match given shape   \\
add              & \textcolor{green}{objset} × \textcolor{purple}{object} → \textcolor{green}{objset}                     & Input set with input object added to it                                                                                                                   \\
remove           & \textcolor{green}{objset} × \textcolor{purple}{object} → \textcolor{green}{objset}                     & Input set with input object removed from it                                                                                                               \\
add\_rel         & \begin{tabular}[c]{@{}l@{}}\textcolor{green}{objset} × \textcolor{purple}{object} x \textcolor{purple}{object} \\ x \textcolor{blue}{relation} → \textcolor{green}{objset}\end{tabular} & \begin{tabular}[c]{@{}l@{}}Input set with new object (first input) added at the \\ given spatial location relative to second input object\end{tabular}    \\
remove\_rel      & \begin{tabular}[c]{@{}l@{}}\textcolor{green}{objset} × \textcolor{purple}{object} x \textcolor{purple}{object} \\ x \textcolor{blue}{relation} → \textcolor{green}{objset}\end{tabular} & \begin{tabular}[c]{@{}l@{}}Input set with object (first input) removed from the  \\ given spatial location relative to second input object\end{tabular}    \\
change\_loc      & \begin{tabular}[c]{@{}l@{}}\textcolor{green}{objset} × \textcolor{purple}{object} x \textcolor{purple}{object} \\ x \textcolor{blue}{relation} → \textcolor{green}{objset}\end{tabular}                & \begin{tabular}[c]{@{}l@{}}Input set with object (first input) location changed to a \\ given spatial location relative to second input object\end{tabular} \\
change\_size     & \textcolor{green}{objset} × \textcolor{blue}{size} → \textcolor{green}{objset}                       & Input set with size updated to the given value                                                                            \\
change\_color    & \textcolor{green}{objset} × \textcolor{blue}{color} → \textcolor{green}{objset}                      & Input set with color updated to the given value                                                                           \\
change\_material & \textcolor{green}{objset} × \textcolor{blue}{material} → \textcolor{green}{objset}                   & Input set with material updated to the given value                                                             \\
change\_shape    & \textcolor{green}{objset} × \textcolor{blue}{shape} → \textcolor{green}{objset}                      & Input set with shape updated to the given value
\\ \bottomrule
\end{tabular}
\caption{(upper) Original function catalog for CLEVR proposed in  \cite{johnson2017clevr}, which we reuse in our data creation process (lower) New functions added to the function catalog for the \texttt{CLEVR\_HYP} dataset.}
\label{tab:fpcat}
\end{table*}
 
  \begin{figure}[h]
 \begin{center}
  \includegraphics[width=0.7\linewidth]{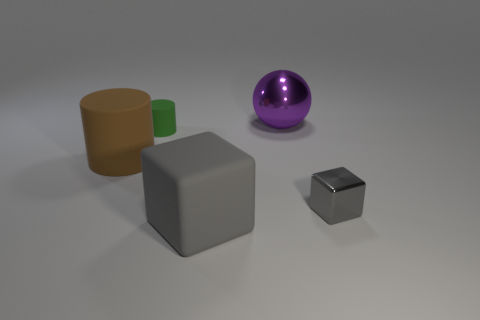} \end{center} 
 \paragraph{\small{small gray metal cube:}} \small{[small gray object, small metal object, small cube, small gray cube, small gray metal object, gray metal cube, small gray metal cube] }
\noindent\paragraph{\small{large brown rubber cylinder:}} \small{[brown object, large brown object, large cylinder, brown rubber object, brown cylinder, large brown rubber object, large brown cylinder,  brown rubber cylinder, large brown rubber cylinder]}
  \caption{Object paraphrases for 2 objects in the scene}
  \label{fig:paraph}
\end{figure}

\paragraph{Object Name Paraphrasing}
There can be many ways an object can be referred in the scene. For example, `large purple metal sphere' in image below can also be referred to as `sphere' as there is no other sphere present in the image. In order to make templates more challenging, we use these alternative expressions to refer objects in the action text or question. We wrote a python script that takes scene graph of the image and generates all possible names one can uniquely refer for each object in the scene. When paraphrasing is performed, one of the generated names is randomly chosen and replaced. Figure 9 demonstrates list of all possible name variants for two objects in the given image.

\paragraph{Synonyms for Paraphrasing}
We use word synonyms file provided with CLEVR dataset generation code. 

\paragraph{Sentence/Question Level Paraphrasing}
For action text paraphrasing, we use Fairseq \cite{ott2019fairseq} based paraphrasing tool which uses round-trip translation and mixture of experts \cite{shen2019mixture}. Specifically, we use pre-trained round-trip models (En-Fr and Fr-En) and choose top-5 paraphrases manually for each template. 
For question paraphrasing, the quality of round-trip translation and mixture of experts was not satisfactory. Therefore, we use Text-To-Text Transfer Transformer (T5) \cite{2020t5} fine-tuned over positive samples from Quora Question Pairs (QQP) dataset \cite{iyer2017first} and choose top-5 per template.   

\subsection{Computational Resources}
All of our experiments are performed over Tesla V100-PCIE-16GB GPU. 

\begin{figure*}
        \begin{minipage}{0.2\linewidth}
            [1] \includegraphics[width=\textwidth,height=0.7\linewidth]{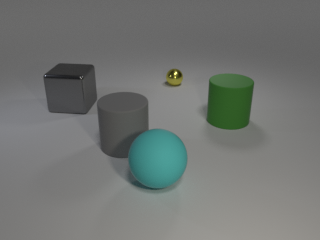}
            \end{minipage} 
            \begin{minipage}{0.70\linewidth}
          
          \small{
       \textbf{T$_A$}: A small red sphere is added to the right of the green object. \\
        \textbf{Q$_H$}: There is a gray cylinder; how many spheres are to the right of it? \\
        \textbf{A}: 2 \\
        \textbf{Classification}: Add action, Counting question \\
        \textbf{Split}: val
             }
             
        \end{minipage}

        \begin{minipage}{0.2\linewidth}
            [2] \includegraphics[width=\textwidth,height=0.7\linewidth]{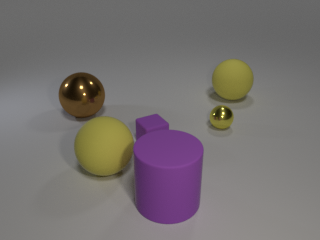}
            \end{minipage} 
            \begin{minipage}{0.70\linewidth}
            
          \small{
       \textbf{T$_A$}: All the purple objects become metallic. \\
        \textbf{Q$_H$}: What number of shiny things are to the left of the small yellow sphere?  \\
        \textbf{A}: 3 \\
        \textbf{Classification}: Change action, Counting question \\
        \textbf{Split}: val
             }
             
        \end{minipage}
        
        \begin{minipage}{0.2\linewidth}
            [3] \includegraphics[width=\textwidth,height=0.7\linewidth]{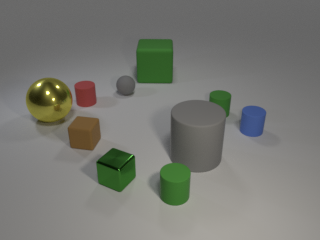}
            \end{minipage} 
            \begin{minipage}{0.70\linewidth}
            
          \small{
       \textbf{T$_A$}: John puts a large red metal cube behind the blue rubber cylinder.  \\
        \textbf{Q$_H$}: There is a small green cylinder that is in front of  the gray thing; are there any large red things behind it? \\
        \textbf{A}: Yes \\
        \textbf{Classification}: Add action, Existence question \\
        \textbf{Split}: val
             }
             
        \end{minipage}
        
        \begin{minipage}{0.2\linewidth}
            [4] \includegraphics[width=\textwidth,height=0.7\linewidth]{clevr_samples/img7.png}
            \end{minipage} 
            \begin{minipage}{0.70\linewidth}
            
          \small{
       \textbf{T$_A$}: Remove all matte objects from the scene. \\
        \textbf{Q$_H$}: Is there any large sphere? \\ 
        \textbf{A}: No \\
        \textbf{Classification}: Remove action, Existence question   \\
        \textbf{Split}: val
             }
             
        \end{minipage}

        \begin{minipage}{0.2\linewidth}
            [5] \includegraphics[width=\textwidth,height=0.7\linewidth]{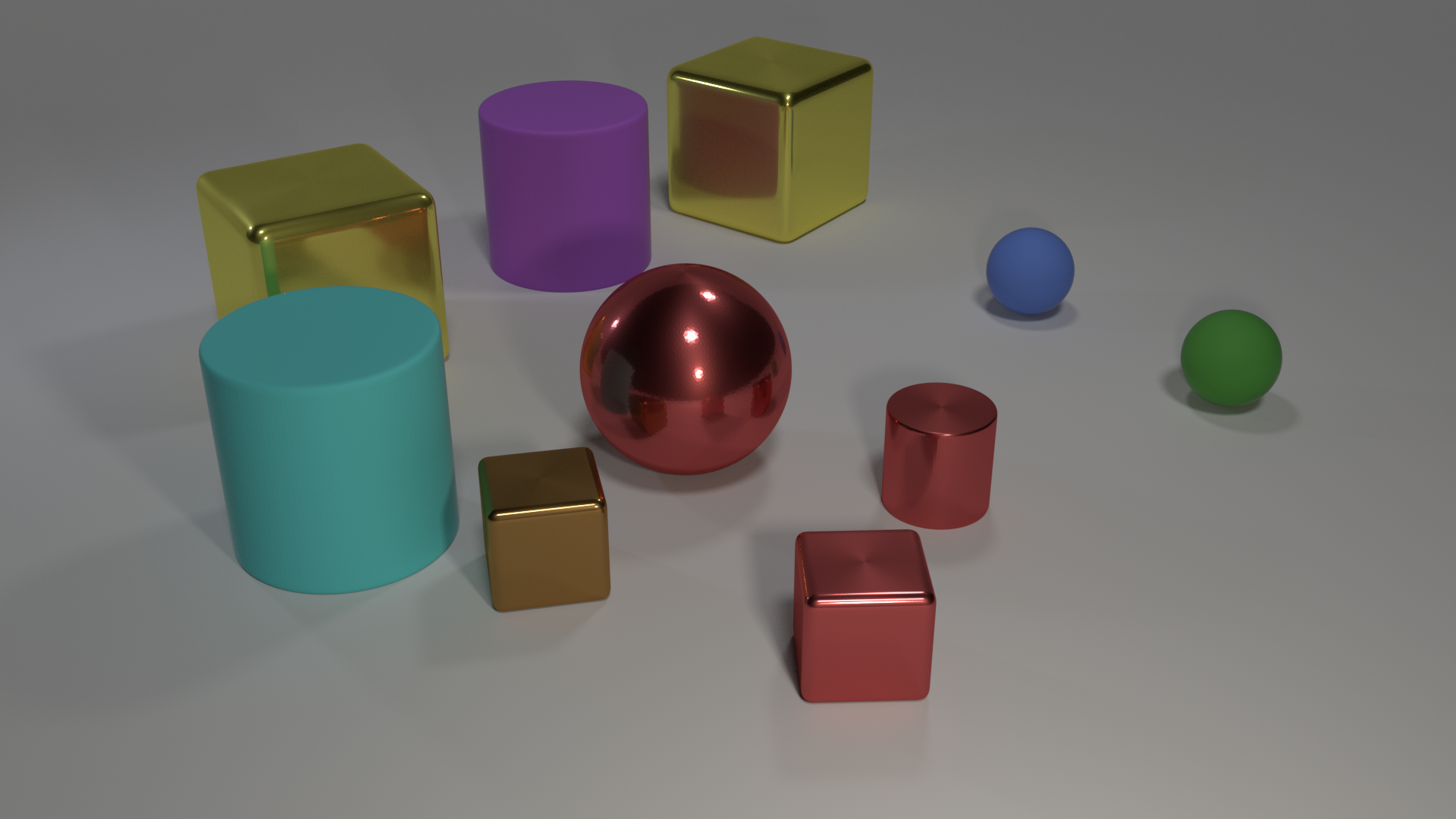}
            \end{minipage} 
            \begin{minipage}{0.70\linewidth}
            
          \small{
       \textbf{T$_A$}: The large cylinder behind the red shiny sphere is moved in front of the green sphere. \\
        \textbf{Q$_H$}: Is there a purple object that is to the right of the big yellow cube that is behind the cyan rubber sphere? \\
        \textbf{A}: No \\
        \textbf{Classification}: Move (in-plane) action, Existence question  \\
        \textbf{Split}: val
             }
             
        \end{minipage}
                    
          \begin{minipage}{0.2\linewidth}
            [6] \includegraphics[width=\textwidth,height=0.7\linewidth]{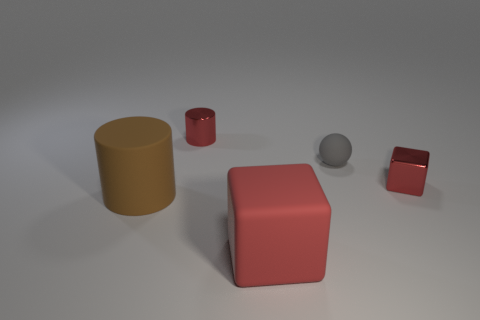}
            \end{minipage} 
            \begin{minipage}{0.70\linewidth}
            
          \small{
       \textbf{T$_A$}: A small green metal sphere is added behind the small red cube. \\
        \textbf{Q$_H$}: What color is the large cylinder that is to the right of the green object? \\
        \textbf{A}: Brown \\
        \textbf{Classification}: Add action, Query Attribute question \\
        \textbf{Split}: val
             }
             
        \end{minipage}
        
        \begin{minipage}{0.2\linewidth}
            [7] \includegraphics[width=\textwidth,height=0.7\linewidth]{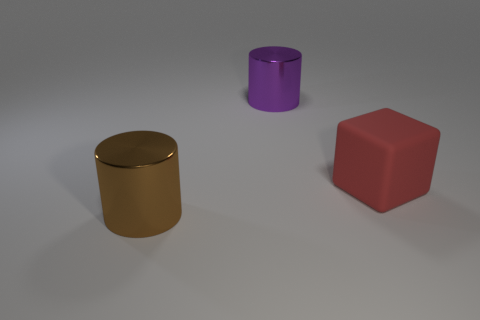}
            \end{minipage} 
            \begin{minipage}{0.70\linewidth}
            
          \small{
       \textbf{T$_A$}: The purple cylinder behind the cube disappers from the scene. \\
        \textbf{Q$_H$}: What material is the object on the left of brown metal cylinder? \\
        \textbf{A}: Rubber \\
        \textbf{Classification}: Remove action, Query Attribute question \\
        \textbf{Split}: val
             }
             
        \end{minipage}
        
           \begin{minipage}{0.2\linewidth}
            [8] \includegraphics[width=\textwidth,height=0.7\linewidth]{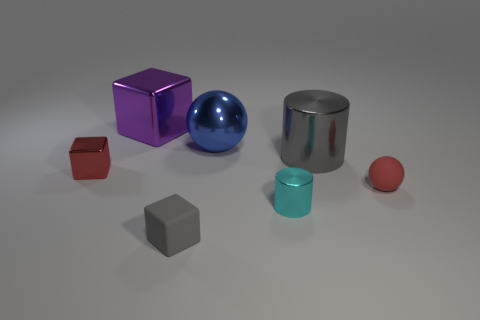}
            \end{minipage} 
            \begin{minipage}{0.70\linewidth}
            
          \small{
       \textbf{T$_A$}: There is a sphere that is to the left of the gray cylinder; it shrinks in size. \\
        \textbf{Q$_H$}:  What size is the blue object? \\
        \textbf{A}: Small \\
        \textbf{Classification}: Change action, Query Attribute question \\
        \textbf{Split}: val
             }
             
        \end{minipage}
        
           \begin{minipage}{0.2\linewidth}
            [9] \includegraphics[width=\textwidth,height=0.7\linewidth]{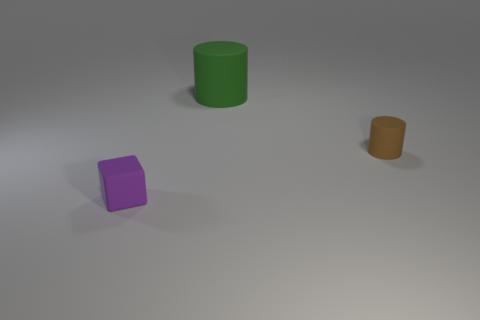}
            \end{minipage} 
            \begin{minipage}{0.70\linewidth}
            
          \small{
       \textbf{T$_A$}: The brown thing is moved in front of the pink rubber cube. \\
        \textbf{Q$_H$}:  What shape is the object that is in front of the pink rubber cube? \\
        \textbf{A}: Cylinder \\
        \textbf{Classification}: Move (in-plane) action, Query Attribute question \\
        \textbf{Split}: val
             }
             
        \end{minipage}

\caption{More examples from the \texttt{CLEVR\_HYP} dataset} 

  \label{fig:moreexamples}
\end{figure*}

\begin{figure*}

        \begin{minipage}{0.2\linewidth}
            [10] \includegraphics[width=\textwidth,height=0.7\linewidth]{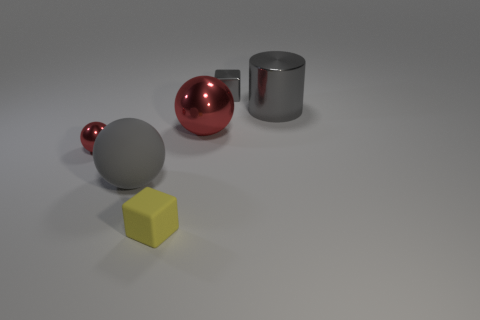}
            \end{minipage} 
            \begin{minipage}{0.70\linewidth}
            
          \small{
       \textbf{T$_A$}: The small red sphere is moved onto the small cube that is in front of the gray sphere. \\
        \textbf{Q$_H$}: What material is the object that is below the small metal sphere? \\
        \textbf{A}: Rubber \\
        \textbf{Classification}: Move (out-of-plane) action, Query Attribute question \\
        \textbf{Split}: val
             }
             
        \end{minipage}
        
        \begin{minipage}{0.2\linewidth}
            [11] \includegraphics[width=\textwidth,height=0.7\linewidth]{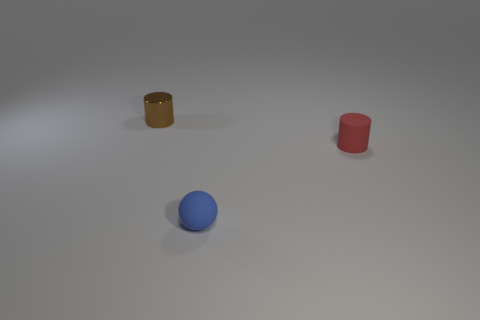}
            \end{minipage} 
            \begin{minipage}{0.70\linewidth}
            
          \small{
       \textbf{T$_A$}: A small yellow metal object is placed to the right of red cylinder; it inherits its shape from the blue object. \\
        \textbf{Q$_H$}: Are there any other things that have the same shape as the blue matte object? \\
        \textbf{A}: Yes \\
        \textbf{Classification}: Add action, Compare Attribute question \\
        \textbf{Split}: val
             }
             
        \end{minipage}
        
        \begin{minipage}{0.2\linewidth}
            [12] \includegraphics[width=\textwidth,height=0.7\linewidth]{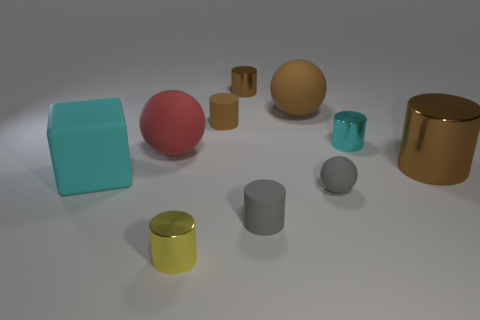}
            \end{minipage} 
            \begin{minipage}{0.70\linewidth}
            
          \small{
       \textbf{T$_A$}: Hide all the cylinders from the scene. \\
        \textbf{Q$_H$}: Are there any other things that have the same size as the gray sphere?
 \\
        \textbf{A}: No \\
        \textbf{Classification}: Remove action, Compare Attribute question \\
        \textbf{Split}: val
             }
             
        \end{minipage}

          \begin{minipage}{0.2\linewidth}
            [13] \includegraphics[width=\textwidth,height=0.7\linewidth]{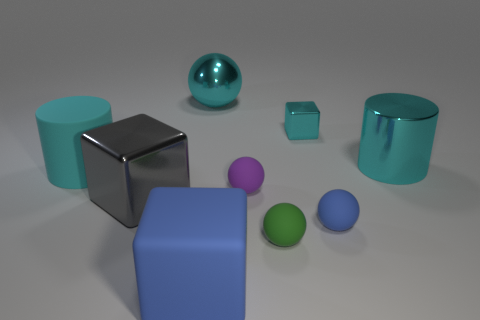}
            \end{minipage} 
            \begin{minipage}{0.70\linewidth}
            
          \small{
       \textbf{T$_A$}: The small block is displaced and put on the left of the blue cube.  \\
        \textbf{Q$_H$}: Is there anything else on the right of the cyan sphere that has the same color as the large metal cylinder?  \\
        \textbf{A}: No \\
        \textbf{Classification}: Move (in-plane) action, Compare Attribute question \\
        \textbf{Split}: val
             }
             
        \end{minipage}
        
        \begin{minipage}{0.2\linewidth}
            [14] \includegraphics[width=\textwidth,height=0.7\linewidth]{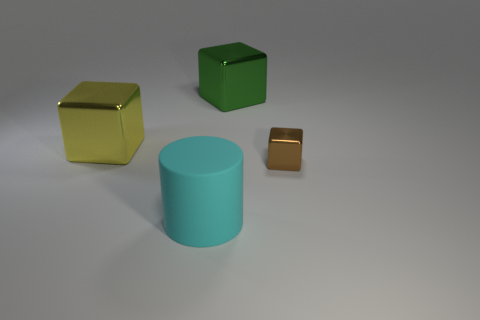}
            \end{minipage} 
            \begin{minipage}{0.70\linewidth}
          
          \small{
       \textbf{T$_A$}: Jill places the small cube on the large cube that is to the left of cyan cylinder. \\
        \textbf{Q$_H$}: There is an object below the brown cube; does it have the same shape as the green object? \\
        \textbf{A}: Yes  \\
        \textbf{Classification}: Move (out-of-plane) action, Compare Attribute question  \\
        \textbf{Split}: val
             }
        
        \end{minipage}
        
        \begin{minipage}{0.2\linewidth}
            [15] \includegraphics[width=\textwidth,height=0.7\linewidth]{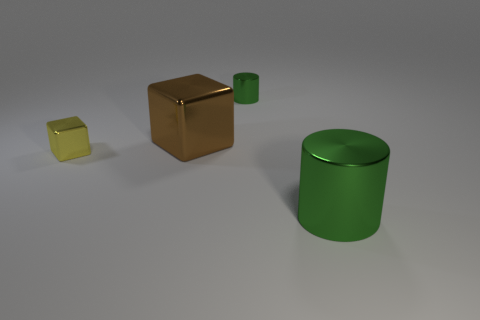}
            \end{minipage} 
            \begin{minipage}{0.70\linewidth}
            
          \small{
       \textbf{T$_A$}: A small brown cube is added to the scene which is made of same material as the golden block. \\
        \textbf{Q$_H$}: Are there an equal number of green objects and brown cubes? \\
        \textbf{A}: Yes \\
        \textbf{Classification}: Add action, Compare Integer question \\
        \textbf{Split}: val
             }
             
        \end{minipage}
        
           \begin{minipage}{0.2\linewidth}
            [16] \includegraphics[width=\textwidth,height=0.7\linewidth]{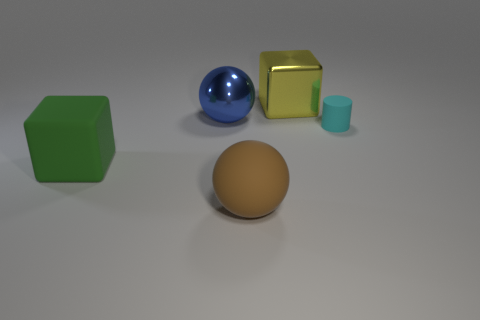}
            \end{minipage} 
            \begin{minipage}{0.70\linewidth}
            
          \small{
       \textbf{T$_A$}: The tiny cylinder is withdrawn from the scene. \\
        \textbf{Q$_H$}: Is the number of rubber objects greater than the number of shiny objects? \\
        \textbf{A}: No \\
        \textbf{Classification}: Remove action, Compare Integer question \\
        \textbf{Split}: val
             }
             
        \end{minipage}
        
           \begin{minipage}{0.2\linewidth}
            [17] \includegraphics[width=\textwidth,height=0.7\linewidth]{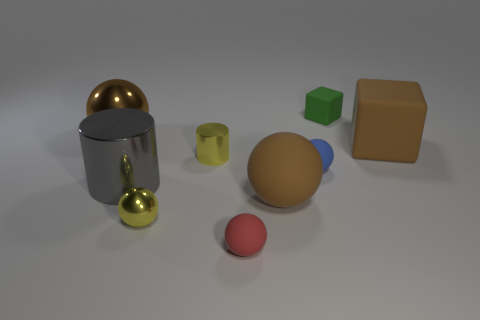}
            \end{minipage} 
            \begin{minipage}{0.70\linewidth}
            
          \small{
       \textbf{T$_A$}: All small metal spheres are transformed into cylinders. \\
        \textbf{Q$_H$}: Are there fewer brown objects that are to the right of the red sphere than the cylinders? \\
        \textbf{A}: Yes \\
        \textbf{Classification}: Change action, Compare Integer question \\
        \textbf{Split}: val
             }
             
        \end{minipage}
        
        \begin{minipage}{0.2\linewidth}
            [18] \includegraphics[width=\textwidth,height=0.7\linewidth]{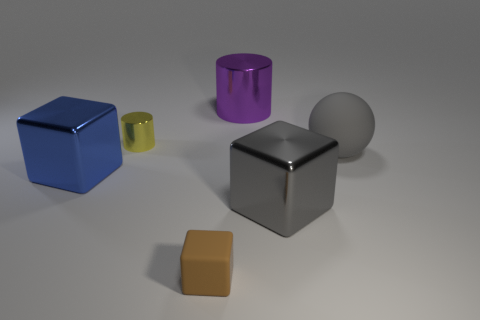}
            \end{minipage} 
            \begin{minipage}{0.70\linewidth}
            
          \small{
       \textbf{T$_A$}: The sphere is placed in front of the large blue cube that is to the left of the yellow shiny object.  
 \\
        \textbf{Q$_H$}: Are there an equal number of gray things to the right of the brown rubber cube and cylinders?
 \\
        \textbf{A}: No\\
        \textbf{Classification}: Move (in-plane) action, Compare Integer question \\
        \textbf{Split}: val
             }
             
        \end{minipage}

\caption{More examples from the \texttt{CLEVR\_HYP} dataset} 

  \label{fig:moreexamples2}
\end{figure*}

\end{document}